\documentclass[letterpaper]{llncs}
\usepackage{times}
\usepackage{helvet}
\usepackage{courier}
\usepackage[hyphens]{url}
\usepackage{graphicx}
\usepackage{subfig}
\usepackage{graphicx}
\usepackage[linesnumbered,lined,boxed,commentsnumbered]{algorithm2e}

\usepackage{xcolor}

\usepackage[htt]{hyphenat}

\newcommand{\sinter}{\mathcal{S}_{\textrm{inter}}}
\newcommand{\snouns}{\mathcal{S}_{\textrm{nouns}}}
\newcommand{\snames}{\mathcal{S}_{\textrm{names}}}
\newcommand{\simage}{\mathcal{S}_{\textrm{image}}}
\newcommand{\scaption}{\mathcal{S}_{\textrm{caption}}}
\newcommand{\prop}{CAPTION}

\title{CAPTION: Correction by Analyses, POS-Tagging and Interpretation of Objects using only Nouns}
\author{Leonardo Anjoletto Ferreira\inst{1} \and Douglas De Rizzo Meneghetti\inst{1} \and \\ Paulo Eduardo Santos\inst{1,2}\\
\institute{Centro Universit\'ario FEI -- S\~ao Bernardo do Campo, S\~ao Paulo -- Brazil \and College of Science and Engineering, Flinders University -- Adelaide -- Australia}
}
\begin{document}
\maketitle
\begin{abstract}
Recently, Deep Learning (DL) methods have shown an excellent performance in image captioning and visual question answering. However, despite their performance, DL methods do not learn the semantics of the words that are being used to describe a scene, making it difficult to spot incorrect words used in captions or to interchange words that have similar meanings. This work proposes a combination of  DL methods for object detection and natural language processing to validate image's captions. We test our method in the FOIL-COCO data set, since it provides correct and incorrect captions for various images using only objects represented in the MS-COCO image data set. Results show that our method has a good overall performance, in some cases similar to the human performance.
\end{abstract}
% -------------------------------------------------- %

\section{Introduction} \label{sec:intro}
Recently, Deep Learning (DL) methods have shown an excellent performance in image captioning and visual question answering. However, it has also been shown that, despite its performance, DL methods do not learn the semantics of the words that are being used to describe a scene, making it difficult to spot incorrect terms used in captions or to substitute words with their synonyms. The method we propose uses natural language processing (NLP) to add meaning to terms used in captions along with a DL method for object recognition, in order to maintain consistency in image captioning. We are going to use the FOIL-COCO \cite{foil} data set as test bed, since it provides both correct and incorrect captions for various images using only objects represented in the MS-COCO image data set \cite{coco}.

 The FOIL-COCO~\cite{foil} data set is a collection of annotations on top of MS-COCO, which provides a caption that can be either correct or have one wrong word. This data set has been proposed to test ML methods ability to comprehend and give meaning to terms used when generating captions, allowing the following three tasks: 1) classify the caption as correct or not; 2) if it is incorrect, find the mistake and 3) fix the caption by replacing the wrong word. Shekhar \textit{et al.} \cite{foil} shows that, although the tested ML methods had a great performance in Visual Question Answering (VQA) tasks \cite{vqa}, it performs poorly in FOIL-COCO. The solution for this problem calls for an appropriate combination of knowledge representation and reasoning methods with deep learning strategies in order to make sense of the captions, connect the words with objects and apply inferences to find the appropriate word to fix the wrong caption.

 In contrast to what is traditionally done in this area, this work uses Deep Learning models not as single method for captioning, but to extract information from the image (e.g., objects, actions, relations) that could be used by a high-level reasoning system. The result of performing NLP in the caption is, then, applied to describe facts about the world and to reason about the image, maintaining consistency of the caption with respect to the objects in the images and the relations depicted. The results of this project will cause a positive impact on advanced sensing and perception, enhancing situation awareness for both users and automated systems, while also promoting rapid understanding of the environment, enabling quick and effective decision making.
 
 The next section presents in more details the VQA problem and the automatic caption generation, along the FOIL-COCO data set. We then describe object detection using DL and the caption processing using NLP (in the Background section), which provides the set of tools needed to develop the \prop{} algorithm, that is presented in the Proposal section. Section Experiments presents the tests executed to evaluate \prop{}, followed by Results and Discussion of the overall performance of this algorithm.
% -------------------------------------------------- %

\section{Related Work} \label{sec:rel}
The goal of object detection is to locate objects pertaining to instances of specific classes in visual inputs, such as images and videos. Although object detection has been a prominent task in the field of computer vision, recent advances in Deep Learning, and especially in Neural Network Architectures specialised in processing visual inputs such as Convolutional Neural Networks (CNN), have paved the way for the creation of new methods which vastly improved the results of existing image classification and object detection challenges \cite{Russakovsky2015}. Furthermore, multiple data sets of annotated images are available online \cite{coco,Russakovsky2015,Kuznetsova2018}, enabling new models to be easily trained, tested and compared under standard conditions.

Recent successful object detection techniques include {\em You Only Look Once} (YOLO) \cite{Redmon2016}, {\em Single Shot MultiBox Detector} (SSD) \cite{Liu2016b} and {\em Faster R-CNN} \cite{Ren2015}.

YOLO \cite{Redmon2016} divides the input image into an \(N \times N\) grid, it then generates \(B\) bounding boxes for each cell and predicts \(C\) class probabilities for each bounding box. Each of the predictions is composed by five values, \(x,y,h,w,k\), where \(x\) and \(y\) represent the coordinates of the the bounding box centre, \(h\) and \(w\) represent the bounding boxes' height and width and \(k\) is a class probability. At training time, each of the \(C\) predictors for a bounding box is specialised in a given class. This specialisation is encoded in a multipart loss function, whose sum squared error is minimised via gradient descent.

SSD \cite{Liu2016b} employs a CNN, called the base network, to generate feature maps, as well as additional convolutional layers of multiple sizes to detect objects in multiple scales. Multiple regions of different scales and aspect ratios are evaluated in the target image in order to accomplish detection. Training is also done via gradient descent and may be boosted by techniques such as hard negative mining and data augmentation strategies. The loss function is a weighted sum of localisation and classification loss.

R-CNN \cite{Girshick2014} employs selective search \cite{uijlings2013selective} to extract a fixed number of 2000 region proposals from an input image. All region proposals are normalised to the same size via warping and used as input to a CNN, which learns a fixed-length feature vector for each one. These vectors are then used as input to several Support Vector Machine (SVM) classifiers, each one trained to classify objects of an specific class.

Fast R-CNN \cite{Girshick2015} improves upon R-CNN by processing all region proposals from an input image in a single forward pass, as well as by replacing multiple SVM classifiers with a single fully-connected, softmax output layer for classification of region proposals.

Lastly, Faster R-CNN \cite{Ren2015} uses the same classification strategy as Fast R-CNN, while introducing a Region Proposal Network (RPN) for object localisation. The RPN uses convolutional filters to produce region proposals represented by \(x,y,h,w,k\), where \(x\) and \(y\) represent the coordinates of a region proposal, \(h\) and \(w\) represent its height and width and \(k\) is an ``objectness'' score. The bounding box predictions of the RPN can be trained through gradient descent. Since both the RPN and Fast R-CNN share convolutional filters, while minimising different loss functions, Ren \textit{et al.}\cite{Ren2015} propose alternating the training of both networks until an acceptable performance is reached.

The use of Neural Networks to answer questions about images has seen great advances in recent years. The end-to-end use of neural networks was shown to achieve a high performance in question answering and caption generation tasks, which fermented the creation of various data sets to further test and develop these ideas.

Johnson \textit{et al.}\cite{clevr} presents CLEVR, a data set of 3D rendered objects along with a set of example questions. The aim of CLEVR is to provide a standard set of objects and questions to be used as benchmark. However, the small set of objects that are available in the images and the artificial scenario created lacked the complexity that artificial neural networks were already capable of coping.

A data set that has been extensively used for object detection is the Microsoft Common Objects in COntext (MS-COCO)\cite{coco} which has over 300,000 images with objects divided into 91 types and 11 super-categories (collections of types). Each image has objects that could be easily recognised by a 4 year old and are presented in their common context and not artificially rendered or modified.

%%% [p:] Tem que tratar do MS-COCO antes de cita-lo aqui!
%%% [l:] adicionado

The Visual Question Answering (VQA) data set~\cite{vqa} has been widely used as a testbed for question-answering systems since it expands the MS-COCO~\cite{coco} data set with more information and abstract scenes. Furthermore, for each image, VQA provides a set of at least three challenging questions to be answered by any AI method.

Although machine learning (ML) methods can be used to solve the problem presented by VQA with high accuracy, two issues became apparent. First, it is unknown how much of the visual information presented in the images is really learned and used by the ML methods to answer the proposed questions, as some ML methods that did not use the image and considered only the question being asked had a good performance answering those questions.
%%%% [p:] Frase sem sentido
%%%% [l:] Alterado para explicar que os métodos só usavam a legenda e descartavam a imagem
Second, it has been shown that the VQA data set is biased towards one type of answer in multiple choice questions (e.g., the same answer for different questions), which explains why ML methods that do not use the input images as source of information are able to answer some questions with great accuracy~\cite{foil}.

%%% [p:] ``biased" em que sentido?? Que images????
%%% [l:] como tinham alguns tipos de bias, coloquei como exemplo apenas um deles

The FOIL-COCO~\cite{foil} data set is a set of annotations on top of the MS-COCO which provides a caption that can be either correct or have one wrong noun. This data set has been proposed to test ML methods ability to comprehend and give meaning to terms used when generating captions and proposes three tasks: 1) classify the caption as correct or not; 2) if it is incorrect, find the mistake and 3) correct the wrong word in the caption. However, Shekhar \textit{et al.}\cite{foil} shows that although the tested ML methods had a great performance in VQA, it performs quite poorly in FOIL-COCO. Shekhar \textit{et al.}\cite{foil2} then extends the FOIL-COCO data set with annotations for other attributes such as adjectives, adverbs and prepositions (i.e., spatial relations).

While FOIL-COCO evaluates how much of the image the neural network really comprehends, Tanti \textit{et al.}\cite{tanti2019} demonstrates that the longer the caption gets, the less relevant the image becomes to the captioning system since the caption generation depends more and more on the prediction of the next word instead of the information from the image. Tanti \textit{et al.}\cite{tanti2019} also shows that objects in the image are the most relevant information used by caption generation systems.

The works of Hendricks \textit{et al.}\cite{gve,hendricks2019} are the ones that relate most to our proposal. Hendricks \textit{et al.} \cite{hendricks2019} proposes Phrase Critic, a caption generator that verifies the relevance of a caption to a given image and, thus, can be used both to generate a caption and to check if a caption is wrong. By using an off-the-shelf localisation model called Visual Genome~\cite{vgenome}, it generates possible descriptions of parts of a picture which are used to ground the caption to the image (i.e., it checks if everything described in the caption also appears in the image). The main difference between Hendricks \textit{et al.} \cite{hendricks2019} and this work is how the caption in grounded in the image. While \cite{hendricks2019} uses a LSTM to generate possible explanations to the image and to compare with the output of Visual Genome, we use NLP to process the caption and DL for image processing (object detection). However, the use of Visual Genome instead of a method for object detection is being studied as an improvement for our current implementation.

This paper uses FOIL-COCO as the data set for the experiments since it has been tested with various caption generating DL methods and can be compared with human performance, but contrary to the ML methods for CLEVR, VQA and also FOIL-COCO, which used only Artificial Neural Networks (ANN) to solve the proposed problem, we combine ANN with natural language processing which can provide some simple notion of semantics to the words used in the caption, making easier to explain the answers to the three tasks proposed in~\cite{foil}.
% -------------------------------------------------- %

%%%[p:] vai usar ontologia ou nao??
%%%[l:] esqueci de mudar essa parte

%%%[p:] Nessa seção vc só tratou de related datasets, mas a seção deria ser sobre ``related work". Quais são os principais trabalhos nessa área? Em que eles sao baseados? Como eles funcionam/resolvem esse problema?

\section{Background}

This section presents the items used in the construction of \prop{}, namely object detection using ANN, NLP and WordNet.

%%% [p:] ``detection" ou ``recognition" ??
%%% [l:] detection por serem objetos genéricos (person). Recognition seria reconhecer quem é a pessoa.

% -------------------------------------------------- %
\subsection{Object detection using Artificial Neural Networks}

Early approaches for object detection involve learning a set of descriptors for objects, such as oriented gradients \cite{hog_patent} or scale-invariant features \cite{Lowe1999}, followed by some form of feature matching in new images to achieve detection. These methods, however, suffer from poor generalisation, scalability and performance.

In order to take advantage of the recent groundbreaking work in image classification, subsequent work focused on generating object proposals (i.e. regions in an image where an object may be contained) followed by conventional image classification techniques \cite{Girshick2014,Girshick2015}. More recently, techniques were proposed that combine object localization and classification in a single regression task, allowing deep learning models such as neural networks with convolutional layers to learn both tasks in a supervised way \cite{Redmon2016,Liu2016b,Ren2015}.

Due to a property known as sparse connectivity \cite{Goodfellow2017}, CNNs have the ability to learn local features from the input data, represented as learnable parameters from convolution filters. The use of the convolution operation preserves the spatial information of the input data. Each convolutional filter is applied to multiple regions of an input image, a property known as parameter sharing \cite{Goodfellow2017} reduces the number of parameters of the model, improving its tractability. Lastly, multiple convolutional layers may be stacked to create more expressive, hierarchical feature representations.

All the methods described in the previous sections are trained via supervised learning, using images with objects annotated by bounding boxes. Each bounding box is described by a tuple \(\langle x,y,h,w,k \rangle\), where \(x\) and \(y\) represent the coordinates of the bounding box centre, \(h\) and \(w\) represent the bounding boxes' height and width and \(k\) is the class of which the object inside the bounding box belongs to.

This work uses the Faster R-CNN \cite{Ren2015} technique trained to detect objects described in the MS-COCO \cite{coco} data set. Thus, given an image from MS-COCO, the method returns the bounding boxes of each object identified, along with its class and a probability that the object belongs to the class.

This alone is not enough to find errors in captions. Thus, we pair this output with Natural Language Processing, as described below.

% -------------------------------------------------- %
\subsection{Natural Language Processing}

In this work, the image captions are manipulated mainly by two Natural Language Processing (NLP) methods: {\em tokenization} and {\em tagging}~\cite{nltk}.

Given a text, tokenizing it is the process of splitting the text into subtexts considering a set of predefined symbol that represents the ending of sentences (e.g., punctuation) or of a single word (e.g., spaces). For example, considering the text ``a man riding a motorcycle", tokenizing it by word would give us the list of words [``a", ``man", ``riding", ``a", ``motorcycle"] which does not alter the order of the text or its meaning, but allows each word to be processed both, independently of the other, or considering other words within a particular text window. This process can be used in two ways when dealing with captions. First, given a caption composed of more than one sentence, it is possible to split each sentence and check for errors independently. Second, given a single sentence, it is possible to process it to obtain more information about its constituting words. One such process that can be done to derive more information from the words present in the caption is {\em tagging}.

Each word that constitutes a given sentence can be classified in lexical categories or parts of speech (POS), such as nouns, verbs, adverbs, etc., thus, POS-tagging (or just tagging) is the process of inferring a lexical category for each word in a sentence. Considering the same text that we used for tokenization, tagging it provides us with the list of tuples [(``a", article), (``man", noun), (``riding", verb), (``a", article), (``motorcycle", noun)] giving a category for each word found.

By tokenizing and POS-tagging a caption, we can filter it for the type of information that we want to process (e.g., nouns, verbs and adjectives) instead of having to deal with the complete sentence and trying to infer a meaning to words that are not important for the task at hand.

Although these two methods provide some information about the structure of the text, their use in caption analysis is dependent on a possible meaning of caption words, which is obtained by using the WordNet \cite{wordnet}.

\subsection{WordNet}

Created in the mid-1980s from the theories of human semantic organisation, WordNet is a large lexical database of words arranged into a semantic network. Words that have the same lexical category and represent the same concept are grouped into sets known as cognitive synonyms or synsets. These synsets are interlinked considering lexical relations and conceptual-semantic notions~\cite{SUMO,wordnet}.

The relations between synsets are most of the time defined by a IS-A relation (hyperonym) but can be also defined in terms of part-whole relation (meronymy) or even as Cross-POS relations, using parts of speech. An important aspect of these relations is that, since synsets form a semantic network, it is possible to navigate it and to calculate similarity between them, using the shortest path from one synset to another.

Based on these concepts, the next section introduces the architecture proposed in this work.
% -------------------------------------------------- %

\section{Correction by Analyses, POS-Tagging and Interpretation of Objects using Nouns (CAPTION)} \label{sec:prop}

One common approach for caption generation is to use Artificial Neural Networks (ANN) from end-to-end, i.e. the input of the ANN is an image and the output is the final caption. However, when using the same approach to validate a caption, the ANN is incapable of recognising eventual mistakes in the use of words, since it has not been shown yet if ANN (or any other deep learning tool) is capable of representing the meaning of words~\cite{foil}. In order to verify the generated image captions, we propose the inclusion of an additional step, which takes an image and its related generated caption, and compares the information extracted from both.

\begin{figure*}[ht!]
    \centering
    \includegraphics[width=\textwidth]{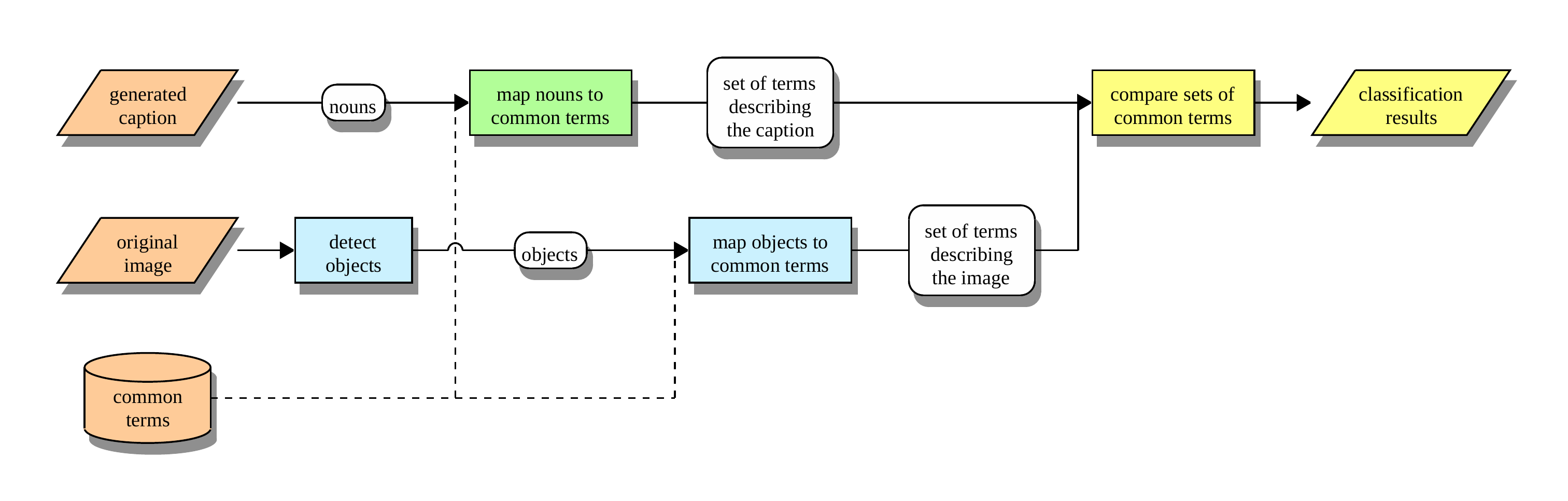}
    \caption{A flowchart of the \prop{} architecture for checking captions}
    \label{fig:proposal}
\end{figure*}

The proposed architecture is presented as a pseudo code in Algorithm~\ref{fig:code} and as a diagram in Figure~\ref{fig:proposal}, which represents the inputs as orange boxes, the steps performed using ANN for object detection are shown in blue boxes, green boxes represent the NLP tasks and yellow boxes are simple set operations used to compare information from the previous steps. Each of these steps are described in details in the following sections.
% -------------------------------------------------- %
\IncMargin{1em}
\begin{algorithm}[ht!]
\SetKwInOut{Input}{Input:}
\SetKwInOut{Output}{Output:}
\Input{an image and a caption}
\Output{foil classification and a dictionary with foil words and corrections}
\BlankLine
$\snames \gets$ detect\_objects(image)\; \BlankLine
tokens $\gets$ tokenize(caption)\; \BlankLine
tags $\gets$ POS-tagging(tokens)\; \BlankLine
$\snouns \gets$ filter\_nouns(tags)\; \BlankLine
\BlankLine
$\sinter \gets \snouns \cap \snames$\; \BlankLine
$\scaption \gets \snouns - \snames$\; \BlankLine
$\simage \gets \snames - \snouns$\; \BlankLine
\BlankLine
Create $corrections$ as a dictionary\; \BlankLine
\ForEach{$foil \in \scaption$}{
  Add $foil$ as a key to $corrections$\; \BlankLine
  \ForEach{$correction \in \simage$}%
  {\If{$foil$ is similar to $correction$}{
    $corrections[foil] \gets correction$
  }
  }
}
\BlankLine
\eIf{$corrections$ is not empty}{
  \Return: $True$, $corrections$\;
  }{
  \Return: $False$}
\caption{\prop{}'s pseudo code}
\label{fig:code}
\end{algorithm}
% -------------------------------------------------- %
% -------------------------------------------------- %
\subsection{Inputs for the architecture}
The proposed architecture uses three pieces of information in order to validate the caption: the caption itself, the related image and a set of common terms used to represent both caption and objects in the image.

While the caption and its related image are the inputs to be analysed, the common terms provide a bridge between words in the caption and object labels obtained by the image processing. For example, if the caption has ``woman" in it but the object detection can only recognise ``person", the set of common terms is going to be used to map the word ``woman" to ``person".

%%% [p:] Nao tinha entendido o paragrafo acima. Mudei um pouco, veja se era isso que vc queria dizer!
%%% [l:] Desse jeito ficou melhor.

This set of terms is used by both object detection and caption processing steps, as described in the following sections.
% -------------------------------------------------- %
\subsection{Object detection}
In this architecture, the object detection steps (blue boxes in Figure~\ref{fig:proposal}) may be any off-the-shelf method (in our current implementation, the Faster R-CNN is used) that is capable of returning a set of objects that can be found in an image, such as the ones presented in the Object Detection using Artificial Neural Networks section. 

%%% [p:] Qual foi usado nesse trabalho?
%%% [l:] Adicionei o que foi usado

Given an image, we use an object detection method to provide a list of objects recognised in it, without the need of information regarding the position of the object in the image or even the degree of confidence of the object detection. This list of objects is then mapped into a set $\snames$ of words using the set of common terms. This $\snames$ of terms recognised in the image is used to analyse the caption along with the set obtained from the caption processing, presented in the next section.

%%% [p:] Pq ``new set"?? de onde ele veio????
%%% [l:] Alterei a frase para mostrar que veio do mapeamento dos objetos para a lista de termos
% -------------------------------------------------- %
\subsection{Caption processing}
The caption processing steps (green boxes in Figure~\ref{fig:proposal}) are responsible for transforming the automatically generated caption into terms that can be compared to the detected objects in the image.

Given the caption, the first step of this process is to recognise only the nouns in it, which is the information that we expect to be related to the objects in the image. This set $\snouns$ is then mapped to the same set of common terms that were used in the objects detected in the image, so that they can be compared in the next step, named {\em Comparison and classification}.
% -------------------------------------------------- %
\subsection{Comparison and classification}
The caption processing step provides a set of nouns defined in terms of the set of common terms ($\snouns$) and the object detection step provides a set of objects also defined in terms of the set of common terms ($\snames$). The classification of the caption into correct or foil (yellow boxes in Figure~\ref{fig:proposal}) can be done simply in terms of standard set operations (i.e., intersection and difference between sets).

%%% [p:] classification of what?
%%% [l:] adicionado o caption classification into correct or not

\begin{figure*}[ht!]
    \centering
    \subfloat[Original image.]{\includegraphics[width=.48\textwidth]{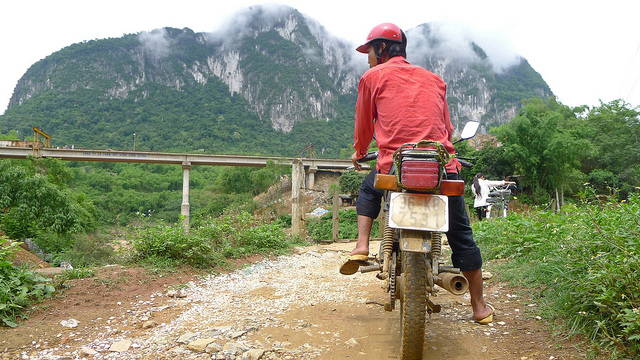}} \hfill
    \subfloat[Marked image.]{\includegraphics[width=.48\textwidth]{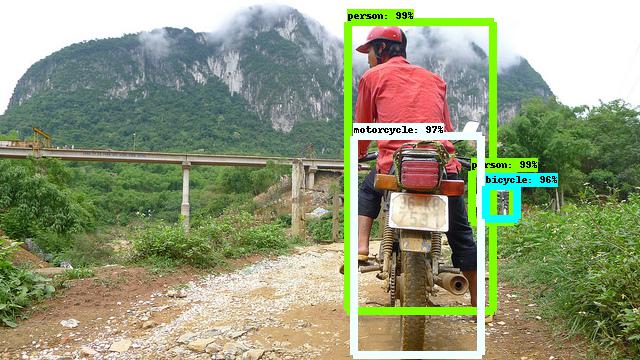}}
    \caption{Example of image that can generate a wrong classification using intersection.}
    \label{fig:ex1}
\end{figure*}

The intersection of both sets ($\sinter = \snouns \cap \snames$) provides information about which objects that were recognised in the image are also in the caption, thus confirming that the object exists. However, we cannot state that the caption is correct only with this information since the wrong word in the caption can be simply another object in the image (e.g., in Figure~\ref{fig:ex1} a man is on a motorcycle and a woman is walking with a bicycle. A wrong caption can state that ``the man is riding a bicycle"). Thus, the intersection tells only which objects were used to generate the caption.

\begin{figure*}[ht!]    
    \subfloat[Original image.]{\includegraphics[width=.48\textwidth]{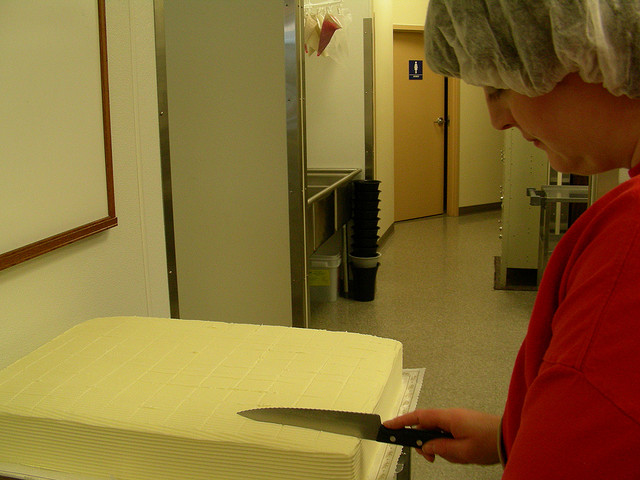}} \hfill
    \subfloat[Marked image.]{\includegraphics[width=.48\textwidth]{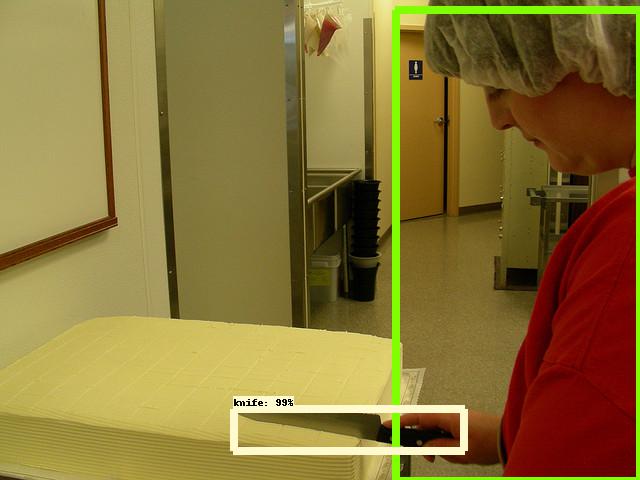}}
    \caption{Example of image that can generate a wrong classification using $\simage$.}
    \label{fig:ex2}
\end{figure*}

The set of objects that are in the image but not in the caption ($\simage = \snames - \snouns$) provides information about which objects should be checked when trying to correct the caption. Once again, we cannot state that the object should be in the caption only because it appears in the image (e.g., in Figure~\ref{fig:ex2}, a woman is using a  knife, recognised in the image, but the caption could only state that ``a woman is cutting a cake" without mentioning the term ``knife").

However, this set $\simage$ can be used to check for possible objects that are potentially wrong in the caption (e.g., if the caption of Figure~\ref{fig:ex2} stated that ``the woman is cutting a pizza" instead of a ``cake") and their properties can be considered in the caption correction (e.g., exchanging ``pizza" for ``cake" in the same example).

The difference between the set of nouns in the caption and the set of objects in the image ($\scaption = \snouns - \snames$) provides information about the objects described in the caption but that were not detected in the image, giving us two possibilities:
\begin{enumerate}
    \item The object detection method failed to find the object in the image and the caption may be correct;
    \item The object described in the caption is not in the image and the caption is wrong.
\end{enumerate}
While the first possibility may be solved by using better object detection methods, the second possibility is exactly the information that we are looking for. If an object described in the caption cannot be found anywhere in the image, we can consider the possibility that the caption is wrong and identify the mistake in it. Therefore, $\scaption$ informs us if there is something in the caption that is not in the image, possibly the wrong word $w_1 \in \scaption$. A possible correction for this caption would then be to change $w_1 \in \scaption$ for a word $w_2 \in \simage$ that can be somewhat equivalent, given their degree of similarity wrt the synsets (e.g., in Figure~\ref{fig:ex2}, ``pizza" and ``cake" are both food and could be exchanged). Therefore, if we are able to find a word $w_1 \in \scaption$ and a word $w_2 \in \simage$ that can substitute $w_1$, the caption can be inferred as wrong and that $w_2$ should substitute $w_1$.

% -------------------------------------------------- %
\subsection{Current implementation}
For this paper, we implemented the architecture described in Figure~\ref{fig:proposal} using a few Python libraries in a container environment.

We used the FOIL-COCO data set~\cite{foil} that provides a set of captions for MS-COCO images~\cite{coco} that may be correct together with the images itself, so that the caption and image would match. Since the captions had more terms than the ones originally described in the images, we used the names and supercategories of MS-COCO as the set of common terms.

The object detection process uses one of the pre-trained TensorFlow~\cite{TF} models available in its model zoo~\cite{Huang2017a}\footnote{The pre-trained models are available at \url{https://github.com/tensorflow/models/tree/master/research/object_detection}}. Since this application does not require to be done in real-time, we employed Faster R-CNN \cite{Liu2016b} with a base network generated through neural architecture search \cite{Zoph2017} on the MS-COCO data set. This choice was made considering the metrics provided by \cite{Huang2017a} for the models available in the model zoo and Faster R-CNN was selected since it provided the best reported mean Average Precision (mAP), despite being the slowest, with a reported 1833 ms for processing a single image. The output of this object detection method is a list of objects already defined in terms of MS-COCO names.

For processing the caption we used NLTK~\cite{nltk} with \texttt{punkt} and \texttt{averaged\_perceptron\_tagger} to, respectively, tokenize and tag words in the caption. NLTK's interface with WordNet~\cite{wordnet} was used to calculate the similarity between nouns found in the caption and the names defined in MS-COCO.

A very important aspect of our current implementation is that we are using only the nouns found in the caption, that were recognised by using NLTK. However, our experiments used the complete FOIL-COCO data set as available in~\cite{foil}.

After detecting objects using Tensorflow and generating the set of nouns using NLTK, the rest of the process used only Python's standard set type operations.

The next section presents the experiments used to test our proposal along with an explanation of how this architecture was implemented in this paper.
% -------------------------------------------------- %
\section{Experiments} \label{sec:xp}
In this section we first present the tasks used to analyse its performance and than the hardware and software setup used during the experiments.

% -------------------------------------------------- %
\subsection{Tasks}
To test the architecture proposed in this work, we used the three tasks described in \cite{foil}:
\begin{enumerate}
    \item Classify the caption as right or wrong (foil);
    \item Detect the wrong word, in case of a foil caption;
    \item Correct the wrong word in the caption.
\end{enumerate}

The three tasks were accomplished by analysing the $\scaption$ and comparing it to $\simage$ so that if $\scaption$ is empty, the caption is considered correct. However, if $\scaption$ is not empty, each of its terms is compared to each of the term in $\simage$ to find a suitable substitution (i.e., if $w_1 \in \scaption$ and $w_2 \in \simage$ have the same MS-COCO supercategory, $w_2$ is a suitable substitute for $w_1$). If a suitable substitute is found, the caption is considered wrong and both, the term considered foil from $\scaption$ and its possible substitute from $\simage$, are informed. However, if a suitable substitute is not found, the caption is considered correct. Thus, in the proposed method, the three tasks are not solved independently since our method finds the foil word (task 2) together with the correction (task 3) which results in classifying the caption as wrong or not (task 1).

Since we used the same data set as~\cite{foil}, the results presented in the next section will be compared with the results provided in that paper for each task.

The next section presents the results obtained by using the proposed method to solve the three tasks presented in this section.
% -------------------------------------------------- %
\subsection{Setup}

The experiments were executed on an Intel Core i7-8700 @ 4.6GHz with 16GB of RAM and a NVidia GeForce GTX 1070 with 8GB of RAM running Debian GNU/Linux Bullseye (the testing version as of this writing). All experiments were run in a container using Docker and NVidia-Docker and an image based on the tensorflow/tensorflow:1.14.0-gpu-py3-jupyter with scikit-learn, opencv-python, nltk and dodo\_detector installed via pip.
% -------------------------------------------------- %
\section{Results} \label{sec:res}
This section presents the results independently for the three tasks presented in the previous section, while using the results presented (and reproduced in this section) by~\cite{foil} and \cite{gve} as baseline for \prop{}. In the latter part of this section, an overall analysis of the performance of the proposed method is presented.
% -------------------------------------------------- %
\subsection{Task 1: Caption classification}

The first task consists of classifying if the caption for a given image is correct or a foil. Table~\ref{tab:res-T1} presents metrics related only to the proposed architecture. It is worth pointing out that \prop{} achieves over 0.7 in precision and recall for both detection of foil caption and in the caption classification as correct.

\begin{table}[ht!]
    \centering
    \begin{tabular}{|l|c|c|c|c|c|}
    \hline
    Caption & Precision & Recall & F1-score \\ \hline
    Correct &   0.81    &  0.74  &   0.77   \\
    Foil    &   0.72    &  0.79  &   0.75   \\ \hline
    \end{tabular}
    \caption{Caption classification results for the proposed architecture.}
    \label{tab:res-T1}
\end{table}

Shekhar \textit{et al.} \cite{foil} performed the same tasks with four different algorithms that used both the caption and the image to classify the caption (CNN + LSTM, IC-Wang, LSTM + norm I and HieCoAtt), and Blind LSTM (a language-only method that did not use any information from the image to classify the caption) and two sets of classifications as done by humans, the first is based on the majority of votes to classify the caption and the second is based on the agreement of all humans judges on the classification. Hendricks \textit{et al.}  \cite{gve} used the Phrase Critic for the same task. Table~\ref{tab:compT1} presents the same results obtained by~\cite{foil} and \cite{gve} along with the ones obtained with \prop{}.

\begin{table}[ht!]
    \centering
    \begin{tabular}{|l|c|c|c|}
    \hline
    Classifier        & Overall(\%) & Correct(\%) & Foil(\%) \\ \hline
    Blind LSTM        & 55.62       & 86.20       & 25.04    \\
    CNN + LSTM        & 61.07       & 89.16       & 32.98    \\
    IC-Wang           & 42.21       & 38.98       & 45.44    \\
    LSTM + norm I     & 63.26       & 92.02       & 34.51    \\
    HieCoAtt          & 64.14       & 91.89       & 36.38    \\ 
    Phrase Critic     & 87.00       &             & 73.72    \\ \hline
    \prop{}           & 76.31       & 80.90       & 71.72    \\ \hline
    Human (majority)  & 92.89       & 91.24       & 92.52    \\
    Human (unanimity) & 76.32       & 73.73       & 78.90    \\
    \hline
    \end{tabular}
    \caption{Results for the classification task presented by~\cite{foil}, \cite{gve} and those obtained by \prop{}. The correct classification results is not provided by \cite{gve}}
    \label{tab:compT1}
\end{table}

Considering only the non-human classifiers presented by \cite{foil}, \prop{} overall performance (76.31\%) is over 10 percentage points above the best (HieCoAtt with 64.14\%). However, when classifying captions as correct, it is only better than IC-Wang (38.98\%), with a performance of about 10 percentage points below the best classifier (LSTM + norm I with 92.02\%). For foil classification, \prop{} performance (71.72\%) is about 25 percentage points higher than the second best (IC-Wang with 45.44\%). Comparing with the classification done by humans, \prop{}'s performance is worse than the classification done by humans voting, but it is surprisingly close to the unanimity classification for the three criteria.

Comparing to the Phrase Critic, its overall performance is about 10 percentage points higher than \prop{} while the foil classification is almost the same (2 percentage points of difference). Although \prop{}'s performance is worst than Phrase Critic, results show how much the ground of the caption to the image is relevant when checking for errors in captions and, most importantly when finding the foil word in the caption, since both Phrase Critic and \prop{} have the top non-human performance in both tasks.

Overall, \prop{}'s performance is good among the three criteria, being close to the human unanimity performance for the same task. \prop{}'s performance is worst only to the Phrase Critic, but again, their approach are much related which may indicate that comparing the information from the image to the caption may be the reason for the good performance.

Given that \prop{}'s classification can only happens when solving the other two tasks, it can be expected that the results for the other two tasks are at least as good as the results for the first one since performing poorly in identifying the error and correcting it would directly affect \prop{}'s performance in classifying the caption as right or foil.
% -------------------------------------------------- %
\subsection{Task 2: Error detection}
Given a foil caption, the second task is identifying which word is wrong in the caption. Although in the FOIL-COCO data set any correct word can be changed to be a foil word, the current implementation of \prop{} considers only nouns as possible foil words. For this task, Shekhar \textit{et al.}\cite{foil} used three algorithms (IC-Wang, LSTM + norm I and HieCoAtt), the same human classification methods (voting and unanimity) and a random classifier (represented by the label 'Chance' in the results). Again, Hendricks \textit{et al.} \cite{gve} used Phrase Critic for this task.

Results for this task are presented in Table~\ref{tab:compT2}. Although \prop{} only consider nouns as possible foil words, we did not compare it to Shekhar \textit{et al.} \cite{foil} methods that also considered only nouns since our method has no \textit{a priori} information about which word is a noun, classifying it during the caption processing steps.

\begin{table}[ht!]
    \centering
    \begin{tabular}{|l|c|c|}
    \hline
    Identifier        & Nouns (\%) & All words (\%) \\ \hline
    Chance            & 23.25      & 15.87          \\
    IC-Wang           & 27.59      & 23.32          \\
    LSTM + norm I     & 26.32      & 24.25          \\
    HieCoAtt          & 38.79      & 33.69          \\ 
    Phrase Critic     &            & 73.72          \\ \hline
    \prop{}           &            & 71.72          \\ \hline
    Human (majority)  &            & 97.00          \\
    Human (unanimity) &            & 73.60          \\
    \hline
    \end{tabular}
    \caption{Results for the error detection task presented by~\cite{foil} along with \prop{}.}
    \label{tab:compT2}
\end{table}

Comparing only the non-human identifiers from \cite{foil}, \prop{}'s performance (71.72\%) is more than two-times better than the second best method (HieCoAtt with 33.69\%). For the human identifiers, we obtained analogous results to those in the first tasks: \prop{}'s performance is worst than voting (97\%) but close to the unanimity (73.60\%).

Comparing to the Phrase Critic, once again our method performance is very close to it, which may not be a surprise considering that both compare information from image and caption to detect the foil word.

Considering the results of this task, and the fact that \prop{} only considers a caption as foil when it detects a foil word and how to correct it, the results for the next task is as important as the results for the first two tasks in the \prop{} overall performance analysis.
% -------------------------------------------------- %
\subsection{Task 3: Error correction}
The last task is to correct the foil word in the caption. In the methods evaluated by \cite{foil}, the caption and the foil word were given  as input to the algorithm. The algorithm's only task was, then, to change the foil word for a correct one. In \prop{}, however, since identifying and correcting the foil word is an important part of the classification step, we did not inform our method which is the foil word for each caption. Instead, we present the results based on the corrections proposed by \prop{} while performing the other two tasks.

For this task, \cite{foil} uses the same non-human methods as the second task (Chance, IC-Wang, LSTM + norm I and HieCoAtt), but this time, no human method was used for comparison. Once more, Phrase Critic was used for this task by \cite{gve}. Results are presented in Table~\ref{tab:compT3} for each of those four methods and also for \prop{}.

\begin{table}[ht!]
    \centering
    \begin{tabular}{|l|c|c|c|}
    \hline
    Method            & All target words (\%) \\ \hline
    Chance            &  1.38                 \\
    IC-Wang           & 22.16                 \\
    LSTM + norm I     &  4.7                  \\
    HieCoAtt          &  4.21                 \\ 
    Phrase Critic     & 49.60                 \\ \hline
    \prop{}           & 90.11                 \\ 
    \hline
    \end{tabular}
    \caption{Results for the word correction task presented by~\cite{foil} along with \prop{}.}
    \label{tab:compT3}
\end{table}

In this task, \prop{}'s performance (90.11\%) is about four times the performance of the best method presented by \cite{foil} (IC-Wang with 22.16\%). Comparing to Phrase Critic, \prop{} performance is superior for a great margin for the first time in the tree tasks. This may be due to the fact that \prop{}'s correction is based on semantics, exchanging words related to each other, while Phrase Critic uses quantitative metrics to indicate possible corrections.

Since performing poorly in this task would interfere with the results of the previous tasks, it was expected for \prop{} to have a good performance in it. We can see that considering the three tasks as a single, larger task with three steps, seems to have helped to develop a method with good overall performance in the three tasks alone.

% -------------------------------------------------- %
\subsection{Discussion}

The first aspect worth pointing out is that \cite{foil} considered the tasks as three separated problems and used distinct, specialised, methods to solve each task. In contrast, \prop{} uses the results of the second and third tasks (error detection and correction) to perform the first task (caption classification), which indicates that, first, it is possible to solve the three tasks at the same time with the same method, without the need to specialise a method for a certain task. Second, information about the second and third tasks is important for solving the first task, since it gives a reason for the caption to be classified as foil or not and, third, combining the three tasks gives us an explainable output. For example, consider Figure~\ref{fig:ex2} and the caption ``a woman cutting a pizza". By combining the three tasks we can tell that the caption is foil (task 1) because there is no pizza in the image (task 2), but the object detection method found a cake and, since cake and pizza are food, the woman may be cutting a cake instead of a pizza (task 3). Thus, \prop{} not only has a better performance than the other methods considered, but also gives a more explainable answer.

%%% [p:] ok, mas isso vc tirou da cartola, o algoritmo nao da' esse tipo de saída (ainda). Certo? I.e. nao é possivel fazer uma query sobre o porque que a caption estava errada. Quando isso for possivel, aí sim o resultado é explicavel...

%%% [l:] isso já está implementado. Para fazer uma query seria necessário passar a imagem e a legenda e o método de classificação retornaria um booleano (T: foil, F: correct) e um dicionário {foild: [correction_1, correction_2, ...]} das palavras encontradas que foram consideradas erradas (foil) e as possíveis substituições ([correction_1, correction_2,...])  

However, there are some setbacks in our method which can be used to explain the results. First, \prop{} depends on the performance of the object detection method to correctly classify the caption. For example, using the same Figure~\ref{fig:ex2} with the same foil caption of ``a woman cutting a pizza", if the object detection method fails to recognise the cake in the image, \prop{} will not be able to find a suitable substitute for pizza and cannot tell that the caption is foil. Thus, it would incorrectly classify the image as correct.

Second, the foil word in the caption can be some other object in the image, which also makes \prop{} to mistakenly classify the image. For example, consider Figure~\ref{fig:ex1} with the caption ``a man riding a bicycle" and an object detection method that detects the man riding the motorcycle and also the woman with the bicycle in the back of the image. In this case, since both objects are in the image and the current implementation of \prop{} only consider the objects but not the relation between them, it would not be able to find the foil word and also classifies the caption as correct.

Third, in the current implementation, a suitable substitution for a foil word is used by finding the most similar object to the foil word that was found, thus \prop{} error correction is as good as the similarity function used to process the objects in the caption and image. As a result, \prop{} can find mistakes in correct captions and also indicate incorrect corrections for foil captions.

Although these three problems are relevant to \prop{}'s performance, improvements in methods for the first and third problems would improve the overall performance of \prop{}. I.e., as objects detection methods improve, the change of not recognising an objects decreases and the first problem becomes harder to occur. For the second problem, a solution is to use more knowledge about the image to confirm the information provided by the caption (e.g., knowing that a man is riding a motorcycle and not a bicycle in the image infers that the caption ``a man riding a bicycle" is wrong).

Lastly, \cite{foil} suggests that the reason for the poor performance of the methods in the three tasks they propose are due to the lack of a word's meaning for a neural network which makes it unable to detect the error and to correct it. Instead of adding meaning to words and objects before being processed by the ANN so that the ANN has information about the meaning of words and may use it to find errors and correct the text, in \prop{} we add meaning to the output of the ANN in a way that do not change what happens inside the ANN. Furthermore, we use a well-defined and accepted corpus (i.e., WordNet) to provide these meanings to words from both image and caption and, since WordNet is constructed as a semantic network, we are capable of performing some operations with those meanings (e.g., calculate similarities). Thus, we direct each method (ANN and NLP) to its most suitable problem (i.e., object detection and text processing, respectively) instead of trying to use a single method to solve the three tasks, as done by \cite{foil} using ANN methods. As a result, we have a method with an overall better performance and also a solution that can be explained.
%%% [p:] não entendi esse ultimo paragrafo!! Está confuso.
%%% [l:] reescrevi o parágrafo para tentar explicar melhor

When comparing to Phrase Critic, \prop{}'s performance is worst for the first task (caption classification), almost identical in the second one (foil word detection) and better in the last one (foil word correction), although the two methods have a lot in common. This shows that comparing (grounding) the caption to the image is an important step for the classification and also provides information for detecting the foil word, indicating that a hybrid approach of combining DL and NLP may have an overall better performance than using just DL methods. However, semantics still seems to play an important role when correcting the caption.

% -------------------------------------------------- %

\section{Conclusion} \label{sec:con}
This paper proposes an architecture to solve the three tasks proposed by \cite{foil} for automatic caption generation for images, which are 1) classify the caption as correct or not, 2) detect the foil word in the caption and 3) correct the caption. 

Our architecture combines object detection of the image with natural language processing of the caption to search for the incorrect word in the caption, suggesting a correction. With this information, the caption is classified as correct or not and the corrections are informed if needed.

Results show that \prop{} has a better overall performance than the methods tested by \cite{foil} and in some tasks it is comparable to human performance. However, there are some drawbacks in our approach such as objects not being recognised by the detector, or the use of poor word similarity functions.

Future works consists in improving the caption processing by using more information, not just nouns, and also considering object relations in the image so that it is possible to overcome some of the shortcomings of the current implementation. These improvements will allow us to test our method in the data set provided by \cite{foil2}.

% -------------------------------------------------- %
\section{Acknowledgements}
Leonardo Anjoletto Ferreira acknowledges that this study was financed in part by the Coordena\c{c}\~{a}o de Aperfei\c{c}oamento de Pessoal de N\'{i}vel Superior -- Brasil (CAPES) -- Finance Code 001. Douglas De Rizzo Meneghetti that this study was financed by the Coordena\c{c}\~{a}o de Aperfei\c{c}oamento de Pessoal de N\'{i}vel Superior -- Brasil (CAPES) -- Finance Code 001.

% -------------------------------------------------- %
\bibliography{bibliography}
\bibliographystyle{splncs04}
\end{document}